\documentclass[letterpaper, 10 pt, conference]{ieeeconf}  

\IEEEoverridecommandlockouts                              

\usepackage{cite}
\usepackage{amsmath,amssymb,amsfonts}
\usepackage{algorithmic}
\usepackage{graphicx}
\usepackage{textcomp}
\usepackage{xcolor}
\usepackage{mathtools}
\usepackage{pgfplots}
\usepackage{filecontents}
\usepackage{tikz}
\usepackage[tight,nice]{units}
\usepackage{subcaption}
\usepackage{rotating}    
\usepackage{makecell}   
\usepackage{tabularray}
\usepackage{adjustbox}
\UseTblrLibrary{booktabs, counter, varwidth}
\usetikzlibrary{spy}
\def\BibTeX{{\rm B\kern-.05em{\sc i\kern-.025em b}\kern-.08em
    T\kern-.1667em\lower.7ex\hbox{E}\kern-.125emX}}

\usepackage{mathtools}
\usepackage{float}

\title{\LARGE \bf Deep Generic Dynamic Object Detection Based on Dynamic Grid Maps}
\author{Rujiao Yan$^{1}$, Linda Schubert$^{2}$, Alexander Kamm$^{1}$, Matthias Komar$^{1}$, Matthias Schreier$^{1}$
  \thanks{$^{1}$R. Yan, A. Kamm, M. Komar, and M. Schreier are with Continental Autonomous Mobility Germany GmbH, Frankfurt a. M., Germany.   
	  {\tt\small \{rujiao.yan, alexander.kamm, matthias.komar, matthias.schreier\}@continental.com}}
  \thanks{$^{2}$L. Schubert is with ADC Automotive Distance Control GmbH, Lindau a.\ B. 
    {\tt\small linda.schubert@continental.com}}}
		
\begin{document}
	\maketitle
	\thispagestyle{empty}
	\pagestyle{empty}
\begin{abstract}
This paper describes a method to detect generic dynamic objects for automated driving. First, a LiDAR-based dynamic grid is generated online. Second, a deep learning-based detector is trained on the dynamic grid to infer the presence of dynamic objects of any type, which is a prerequisite for safe automated vehicles in arbitrary, edge-case scenarios. The Rotation-equivariant Detector (ReDet)~--~originally designed for oriented object detection on aerial images~--~was chosen due to its high detection performance. Experiments are conducted based on real sensor data and the benefits in comparison to classic dynamic cell clustering strategies are highlighted. The false positive object detection rate is strongly reduced by the proposed approach.
\end{abstract}


\section{Introduction}

\begin{figure}[htbp]
\includegraphics[width=1.00\linewidth]{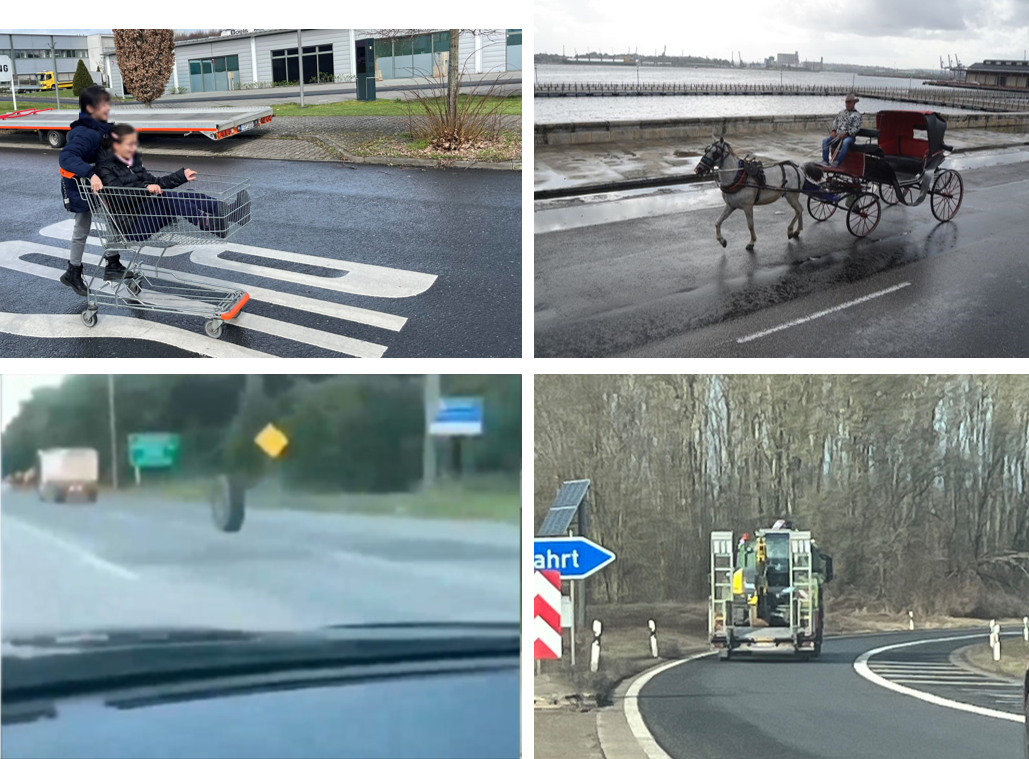}
\caption{Generic dynamic object examples. Object detection methods trained for standard classes are likely to struggle in such scenarios.}
\label{fig:GenericDynamicObjectExamples}
\end{figure}

The perception and representation of the environment is a key ingredient in automated driving systems. A multitude of data fusion methods and ways of modeling the local area around the ego vehicle have been proposed~\cite{Schreier2018,Schreier2022}. With regard to \textit{dynamic} objects, the majority of work focuses on the detection of \textit{known} object classes such as vehicles, cyclists, or pedestrians. Deep neural networks are trained on established labeled datasets such as KITTI or nuScenes based on camera, LiDAR, and RADAR data to detect such predefined object classes~\cite{Feng2020,Mao2023}. In reality, however, the spectrum of objects that can be dynamic is not limited to predefined classes, but nearly anything can move. Examples include shopping carts, rolling tires, or all kinds of animals. But also standard classes such as vehicles exist in all kinds of non-standard appearances, see Fig.~\ref{fig:GenericDynamicObjectExamples}. 
Detectors trained on predefined object classes are incapable to perceive such \textit{generic} dynamic objects~--~let alone to estimate their velocities or accelerations, which can lead to dangerous situations. 

To cope with this problem, so-called dynamic grid maps were introduced~\cite{Nuss2018,Steyer2018,Vatavu2020}, which do not make assumptions about the type or shape of dynamic objects and can estimate a full velocity vector distribution for each grid cell around the ego vehicle via particle filtering alongside arbitrary static environment structures. Dynamic objects are then detected/extracted from such dynamic grids via clustering techniques like~\mbox{DBSCAN}~\cite{Gies2018,Steyer2020}. Alternatively, multiple subsequent static grid maps are post-processed to create, track, and classify generic dynamic object hypotheses~\mbox{\cite{Schreier2014,SchreierIEEETrans1}}. Such hand-designed clustering and classification approaches are, however, subject to false positive detections since dynamic cells in dynamic grids can also emerge from swaying trees in the wind or from newly appearing static environment structures, which are hard to separate from true object motion. Therefore, various deep learning-based approaches were proposed to improve the grid-based dynamic object detection. 

One of the early works of refining the separation of dynamic and static entities in dynamic grids can be found in~\cite{Piewak2017}, in which a fully convolutional network is trained to infer, whether individual dynamic grid cells are moving or not. The result is a cell-wise classification of the surroundings into the classes dynamic and static. No explicit dynamic object detections are outputted from the network. Object representations are, however, beneficial for subsequent tasks in the automated driving stack such as situation interpretation, prediction, and planning. Therefore, a single-stage deep convolutional network was trained in~\cite{Hoermann2018} to directly detect dynamic object hypotheses from dynamic grids consisting of object shape, position, orientation, and an existence score. Promising results were shown on an exemplary urban junction for a stationary ego vehicle. In a follow-up work~\cite{Engel2018}, a single-stage deep convolutional neural network was combined with a recurrent LSTM neural network taking dynamic occupancy grid maps as input and generating dynamic object bounding boxes as a result. The recurrent network is supposed to capture long-term sequential relations, e.g. to overcome occlusions. Results were likewise shown for a static, parked ego vehicle in an urban scenario. In~\cite{Wirges2020}, a single-stage, real-time capable RetinaNet detection network was trained on static grid maps to extract bounding boxes. Since no dynamic grid is used in this work, bounding boxes are also generated for all box-like static objects. This makes subsequent tracking less robust and the interpretation of the traffic scene more complex. 

\begin{figure}
	\centering
		\includegraphics[width=1.00\linewidth]{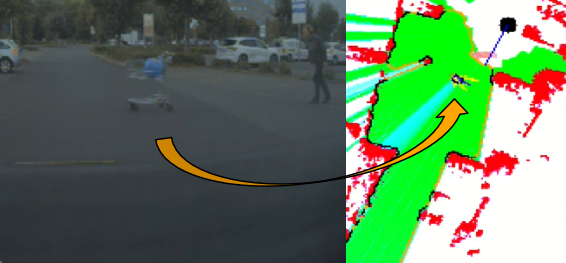}
	\caption{Moving shopping cart detection. Camera image (left) and dynamic grid with overlaid detection results (right).}
	\label{fig:DynGridWithShoppingCart}
\end{figure}

In this paper, we similarly propose to replace the classic cell clustering by a deep learning-based object detection method operating on dynamic grids, which is optionally followed by a high-level object tracker. Our approach is inspired by the mentioned works and extends them to realize~\textit{generic dynamic object detection} in an~\textit{online} fashion from a~\textit{moving ego vehicle}. The Rotation-equivariant Detector~\cite{Han2021} (ReDet)~--~originally designed for oriented object detection on aerial images~--~was chosen to perform the task because of its high detection performance. The grids are treated as multi-channel images and due to the networks' capabilities to make use of context information in the grid, the number of false positives are strongly reduced. In contrast to fully end-to-end trained dynamic grid maps~\cite{Schreiber2022} or deep tracking approaches~\cite{Dequaire2018}, a remarkably low amount of training data is necessary to achieve promising results across a large variety of standard and non-standard dynamic object scenarios. Fig.~\ref{fig:DynGridWithShoppingCart} gives an initial impression of detecting a moving shopping cart, which was never part of the training data. 

After presenting our proposed method in Section~\ref{sec:method} and highlighting details on the experimental setup in Section~\ref{sec:Experimental_Setup}, the main results are discussed in Section~\ref{sec:Results_and_Discussion} before finally summarizing our main contributions in Section~\ref{sec:conclusion}.

\section{Method}\label{sec:method}
We consider the generic dynamic object detection task as a rotated bounding box object detection problem~\cite{Zhou_2022} and treat dynamic grid maps as bird's eye view images. Unlike the majority of camera image-based object detection networks, which output horizontally aligned bounding boxes, rotated object detection, in addition, involves the prediction of the bounding box angle. Rotated object detection networks are normally applied on aerial images, e.g. to detect arbitrarily-rotated vehicles or ships. 

We choose the Rotation-equivariant Detector~\cite{Han2021} (ReDet) due to its high detection performance~\cite{Zhou_2022}. A dynamic occupancy grid map as explained in~\cite{Steyer2018} is applied as input for the detection network, which generates orientated bounding boxes of generic dynamic objects as a result.

\subsection{Dynamic Grid Map as Network Input}\label{sec:input}
We employ the dynamic grid map algorithm presented in~\cite{Steyer2018}. Each grid cell contains Dempster-Shafer basic belief masses $m$ for each element, i.e. hypothesis $\theta$, of the power set  
	\begin{equation}
		2^{\Theta} = \{\emptyset, \{F\}, \{S\}, \{D\}, \{S,D\}, \{F,D\}, \{F,S\}, \Theta\}
	\end{equation}
of a chosen frame of discernment $\Theta=\{F, D, S\}$, so that the sum of all masses equals to one. Here, $\{F\}$ refers to currently free areas, $\{S\}$ to statically occupied areas, $\{D\}$ to currently dynamically occupied areas, $\{S,D\}$ to currently occupied areas (statically or dynamically occupied), $\{F,D\}$ to passable areas, which are either free or dynamically occupied, and $\Theta$ to unknown areas, i.e. areas, which are either free or occupied. The case that a cell is both free and statically occupied is excluded as it is conflicting by definition, therefore~$\{F,S\}$ is omitted. 

For the visualization of such dynamic grid maps, we use the same color coding scheme as introduced in~\cite{Steyer2018}, i.e.
\begin{equation}
		\text{RGB} =\bigg(1-\sum_{{\{S\}}\cap\theta=\emptyset}{m_{\theta}}, 
		1-\sum_{{\{F\}}\cap\theta=\emptyset}{m_{\theta}},
		1-\sum_{{\{D\}}\cap\theta=\emptyset}{m_{\theta}}\bigg)\label{eq:1}
\end{equation}	
for all hypotheses $\theta\subset\Theta$. As a consequence, static occupancy~$\{S\}$ results in red~(R), free space~$\{F\}$ in green~(G), dynamic occupancy~$\{D\}$ in blue~(B), unclassified occupancy~$\{S,D\}$ in magenta, passable areas~$\{F, D\}$ in cyan, and unknown areas~$\{\Theta\}$ in white, see Fig.~\ref{fig:DynGridWithShoppingCart}.

We use the same color coding to encode a dynamic grid map into three channels to obtain a regular RGB image as input for the neural network. Optionally, two additional channels are added by using particle information available in the dynamic grid. The particles approximate the velocity distribution in each grid cell. The two additional channels are formed by normalizing the mean velocity components $v_{x}$ and $v_{y}$ of all particles of a cell.

\subsection{Neural Network Model}

According to~\cite{Zhou_2022}, ReDet~\cite{Han2021} with multiple scale image splits and random rotations outperforms most other rotated object detection methods. Therefore, it is selected for our task. Regular CNNs are translation-equivariant but not rotation-equivariant and consequently require a lot of rotation-augmented data to train an accurate detector for arbitrarily rotated objects. In contrast, ReDet produces rotation-equivariant features in the backbone, which significantly reduces the complexity in modeling orientation variations. In the subsequent detection head, Rotation-invariant Region of Interest Align (RiRoI Align) steps extract instance-level, rotation-invariant features from rotation-equivariant features. For an instance-level, rotation-invariant feature, the output remains identical for any rotational transformation applied to an object. In this work, the rotated box is defined by its center position~$x_{\text{center}}, y_{\text{center}}$, width~$w$, height~$h$, and angle~$\psi$ between the width of the box and the positive $x$-axis with $\psi \in [-90^\circ, 90^\circ]$ and $w>h$. 

\section{Experimental Setup}\label{sec:Experimental_Setup}

\subsection{Baseline: Classic Clustering Method}\label{sec:Base_line}
The classic cell clustering method DBSCAN for extracting dynamic objects is used as a baseline. Only grid cells with a dynamic occupancy mass above a minimum threshold $m_{D,\text{min}}$ are considered for DBSCAN clustering. Its parameters~--~the maximum distance between cluster points $\epsilon_{d}$, the maximum velocity difference $\epsilon_{v}$, and the minimum number of cells for a cluster to be valid $\epsilon_{n}$~--~are fine-tuned manually. Based on experiments, the parameters are set as~\mbox{$m_{D,\text{min}}=0.5$}, \mbox{$\epsilon_{d}=\unit[1.5]{m}$}, \mbox{$\epsilon_{v}=\unit[3]{m/s}$}, and \mbox{$\epsilon_{n}=4$}.  

The method is easy to implement and runs fast. However, it only considers dynamic grid cells and leads to false positives when dynamic occupancy masses are wrong due to incorrect cell velocity estimations, e.g. at guardrails or in case of swaying trees in the wind next to the street. In contrast, deep learning-based object detectors take the spatial scene context contained in the the full grid into account, which leads to improved detection results as later shown in Section~\ref{sec:Results_and_Discussion}.

\subsection{Generation of Ground Truth Data}\label{sec:DataGeneration}
We use a roof-mounted VLS-128 LiDAR with an update rate of \unit[10]{Hz} to generate dynamic grid maps in real-time online in the vehicle. The data is collected from real-world highway and urban driving scenarios. Different locations were chosen to collect training and test data. To reduce the amount of similar training data, only every $5$-th dynamic grid map is used in the labeling process. 

Our ground truth data consists of 3 parts~--~subsequently termed data~1,~2, and~3. \textbf{Data~1} is manually labeled data. Since annotation is time-consuming, only $1450$ frames are hand-labeled ($858$ frames for training, $287$ for validation and $305$ for testing). For \textbf{data~2}, we run the classic DBSCAN approach, see Section~\ref{sec:Base_line}, to auto-label dynamic objects with a manual post-processing to remove frames with false or inaccurate auto-labels. 
This part has $3964$ frames ($3295$ for training, $313$ for validation and $356$ for testing). For \textbf{data~3}, we drove at night through empty streets without any dynamic objects, so that no annotation is required. It contains $1171$ frames ($603$ for training, $100$ for validation and $100$ for testing). These frames are used as negative examples for the neural network to better learn to suppress false positives. In total, we have $5124$ frames for training, $700$ for validation and $795$ frames for inference. These three subsets are summarized in TABLE~\ref{table:datasets}. 

\begin{table}[h]
		\caption{Data Subsets}
		\label{table:datasets}
		\begin{center}
			\begin{tabular}{l||c|c|c|c|c}
				\hline
				 Subset & Training& Validation& Test& Total &Remarks\\
				\hline
				Data~1 &858 &287 &305 & 1450 &Manually labeled\\
				\hline
				Data~2 &3295 &313 &356 & 3964 & DBSCAN\\
				\hline
				Data~3 &971 &100 &100 & 1171 &No dyn. objects \\
				\hline
				Total &5124 &761 &700 &6585&\\
				\hline
			\end{tabular}
		\end{center}
	\end{table} 
	
\subsection{Implementation Details}
We use the pretrained model based on DOTA v1.0~\cite{Xia2018} for $12$ epochs, ReResNet50 as backbone, and SGD optimizer with an initial learning rate of $0.00025$. The learning rate is reduced by factor $10$ at each decay step at $8$ and at $11$ epochs. Note that the learning rate is set very small because the training runs based on the pretrained model. Momentum and weight decay are chosen as $0.9$ and $0.0001$, respectively. All models were trained for $20$ epochs with a batch size of $4$ using an IoU threshold of $0.5$, which ensures network training convergence without overfitting in all cases. The dynamic grid map input has $500 \times 500$ cells with a cell resolution of $\unit[0.2]{m} \times \unit[0.2]{m}$. A single NVIDIA TITAN V was used for training and inference.
 
\section{Results and Discussion}\label{sec:Results_and_Discussion}

In the first results Section~\ref{sec:different_datasets}, we verify that the extension of the training dataset by data without manual labeling indeed improves the inference performance. For this purpose, we compare the model ReDet trained with combinations of different datasets. Subsequently, we evaluate in Section~\ref{sec:different_model_inputs} if it is beneficial to use the potentially redundant \textit{dynamic} information contained in dynamic grids, i.e. dynamic cell masses \textit{and} particle velocity information. To do so, we encode the network's dynamic grid inputs to either 3 channels $\{R,G,B\}$ or to 5 channels $\{R,G,B,v_{x},v_{y}\}$ with normalized mean velocity components~$v_{x}$ and~$v_{y}$ of all particles in a cell. After the best dataset and input channel option are fixed, ReDet is compared with RetinaNet in Section~\ref{sec:RetinaNetComparison}. The latter was trained to detect bounding boxes of predefined object types on static grid maps in~\cite{Lin2017}. Finally, the quantitative and qualitative comparison between ReDet and the classic clustering method DBSCAN is presented in Section~\ref{sec:Comparison_classic}.

\subsection{Comparison of Different Training Datasets}\label{sec:different_datasets}

We trained ReDet with combinations of different datasets previously described in Section~\ref{sec:DataGeneration}. The inference quality (always evaluated on the whole test set from data 1, 2, and 3) is compared in terms of mean Average Precision~mAP, see Table~\ref{table:ComparisonReDet}. 
\begin{table}[htbp]
		\caption{Comparison of ReDet trained on different datasets.}
		\label{table:ComparisonReDet}
		\begin{center}
			\begin{tabular}{l||c|c|c}
				\hline
				Training Datasets & Data 1 & Data 1 + 3 & Data 1 + 2 + 3\\
				\hline
				mAP ($\%$) &56.8 &65.2 &81.2 \\
				\hline
			\end{tabular}
		\end{center}
	\end{table}
	
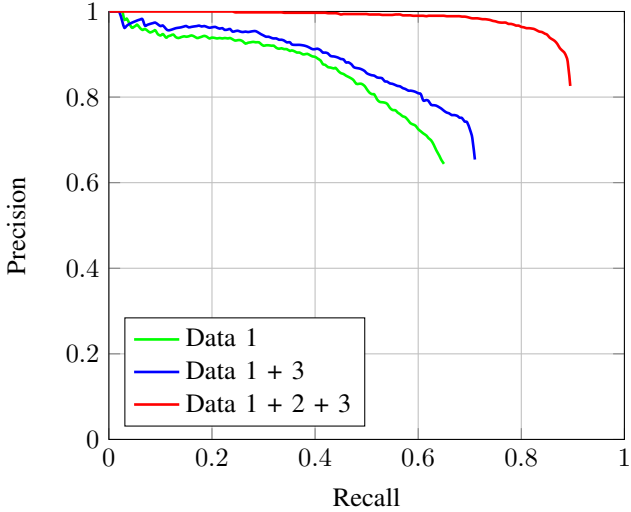
\begin{figure}\centering
\begin{tikzpicture}
    \begin{axis}[
      xlabel={Recall},
      ylabel={Precision},
      legend pos=south west,
      grid=both,
      xmin=0,   
      xmax=1,   %
      ymin=0,   %
      ymax=1,   %
			legend cell align={left}
      ]
      \foreach \modelFile/\mycolor/\modelName in 
      {240118-train-SG-rbg-redet_1256_20.csv/green/data1,
			240128-SG-NoDyn-redet_1256_20.csv/blue/data1+data3,
			240115-full-dataset-rbg-redet_1256_20.csv/red/data1+data2+data3
      }
      {
        \edef
        \temp
        {
            \noexpand
            \addplot[smooth,line width=1pt,color={\mycolor}] table [x=recall, y=precision, col sep=comma] {\modelFile};
        }
        \temp
				\addlegendentryexpanded{Data 1}
				\addlegendentryexpanded{Data 1 + 3}
				\addlegendentryexpanded{Data 1 + 2 + 3}
      } 
    \end{axis}
\end{tikzpicture}
\caption{Precision/Recall curves of ReDet trained on different datasets. The evaluation is always implemented on the whole test set from data 1, 2, and 3.}
\label{fig:PRCscompare}	
\end{figure}

The corresponding precision/recall curves are visualized in Fig.~\ref{fig:PRCscompare}.

It becomes evident that a significant detection performance improvement results from adding data subsets 2 and 3, which both do not require manual labeling. Therefore, the whole training set (data 1 + 2 + 3) is always applied in the following experiments. 

\subsection{Comparison of Different Model Inputs}\label{sec:different_model_inputs}

Since our target is to detect dynamic objects, dynamic information should be used in the input. As described in Section~\ref{sec:input}, we can encode a dynamic grid map into an RGB image so that its dynamic occupancy mass information is already indirectly considered by the cell color. The mean velocities in $x$- and~$y$-direction of all particles in a cell, $v_{x}$ and $v_{y}$, are thus~--~on first sight~--~redundant dynamic information. To see if explicit speed information can further contribute to the detection performance, we encode the dynamic grid map into a $5$ channel matrix with channels $\{R,G,B,v_{x},v_{y}\}$. The results are summarized in Table~\ref{table:2}.

\begin{table}[htbp]
		\caption{Comparison of ReDet with different inputs.}
		\label{table:2}
		\begin{center}
			\begin{tabular}{l||c|c}
				\hline
				Input & $\{R,G,B\}$ & $\{R,G,B,v_{x},v_{y}\}$\\
				\hline
				mAP ($\%$) &81.2 &80.9  \\
				1/Inference Time (fps)&2.6 &2.5
			\end{tabular}
		\end{center}
	\end{table}  
	
It becomes obvious that mAP and inference times are similar and that adding the two velocity channels does thus not result in performance improvements. Consequently, using the 3 RGB channels seems sufficient for the task at hand. 

\subsection{Comparison of ReDet with RetinaNet}\label{sec:RetinaNetComparison}

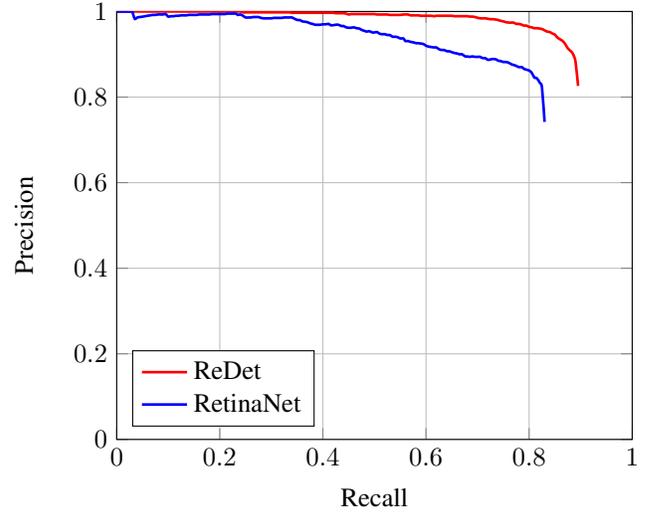
\begin{figure}\centering
\begin{tikzpicture}
    \begin{axis}[
      xlabel={Recall},
      ylabel={Precision},
      legend pos=south west,
      grid=both,
      xmin=0,   
      xmax=1,   %
      ymin=0,   %
      ymax=1,   %
			legend cell align={left}
      ]
      \foreach \modelFile/\mycolor/\modelName in 
      {240115-full-dataset-rbg-redet_1256_20.csv/red/ReDet,
      240123-fulldata-rotated-retina-rbg_1256_20.csv/blue/RetinaNet}
      {
        \edef
        \temp
        {
            \noexpand
            \addplot[smooth,line width=1pt,color={\mycolor}] table [x=recall, y=precision, col sep=comma] {\modelFile};
        }
        \temp
        \addlegendentryexpanded{\modelName}
      } 
    \end{axis}
\end{tikzpicture}
\caption{Precision/Recall curves of ReDet compared to RetinaNet.}
\label{fig:PRCscompareRetina}	
\end{figure}

In~\cite{Wirges2020}, RetinaNet~\cite{Lin2017} was trained to detect bounding boxes of some predefined object types on static grid maps. We evaluate how it compares to ReDet on our dynamic grid maps. To compare the performance in a fair manner, RetinaNet was also trained on the same training set for 20 epochs including multiple image splits and random rotations and using the pretrained model based on DOTA v1.0 for 12 epochs as well. The mAP of RetinaNet based on the whole test set is worse ($78.4\%$) than that of ReDet ($81.2\%$) while the inference frequency of $5.5$ fps is better. The respective precision/recall curves are shown in Fig.~\ref{fig:PRCscompareRetina}. 

\subsection{Comparison with Classic Clustering}\label{sec:Comparison_classic}

\begin{figure}[htbp]
\includegraphics[width=1\linewidth]{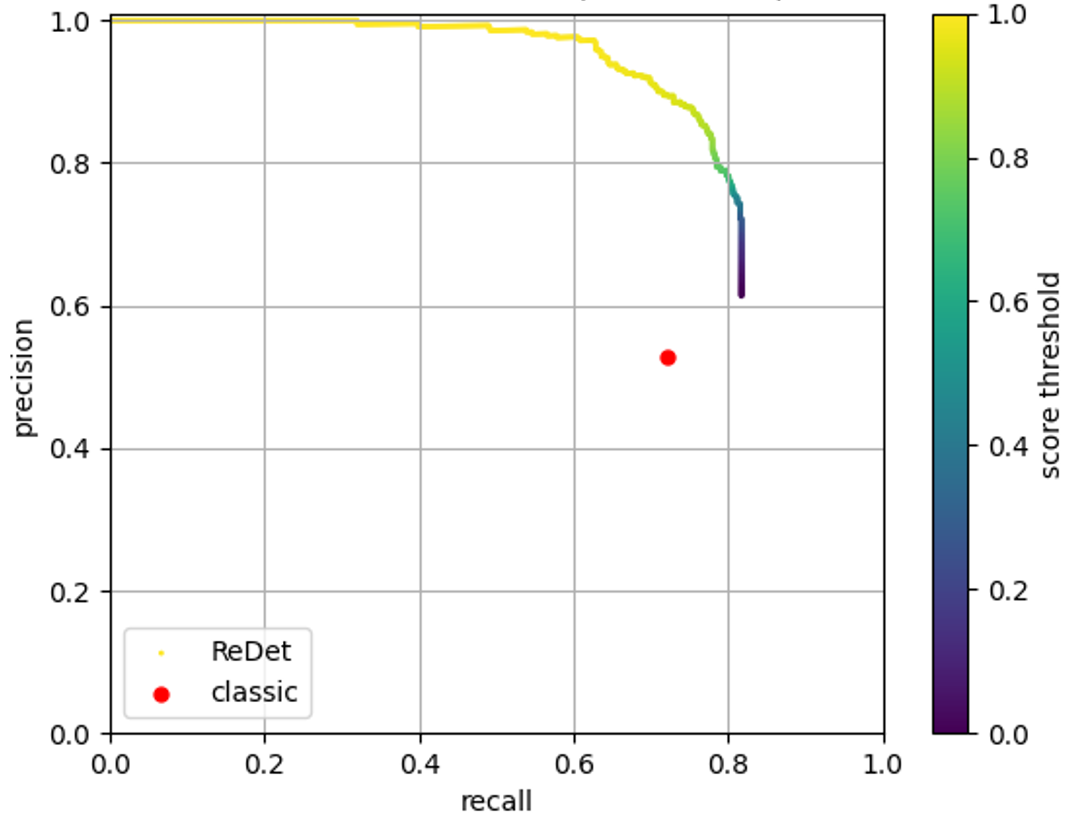}
\caption{ReDet object detection precision/recall curve compared to precision/recall of classic method shown as red dot. ReDet is trained on the whole training set. Both ReDet and the classic DBSCAN are evaluated on data 1 and 3.}
\label{fig:PRCCompareToClassic}
\end{figure}

We now compare the deep learning model with the classic DBSCAN method. To compare both in a fair manner, we exclude data 2 from the test set as its annotations are generated by the classic method. Fig.~\ref{fig:PRCCompareToClassic} shows the ReDet object detection precision/recall curve and the precision/recall obtained from the classic method as a red dot. 

\begin{figure*}[h!]
		\centering
            {\setlength{\fboxsep}{0pt}%
			\setlength{\fboxrule}{0pt}%
			\framebox{\includegraphics[width=1\textwidth]{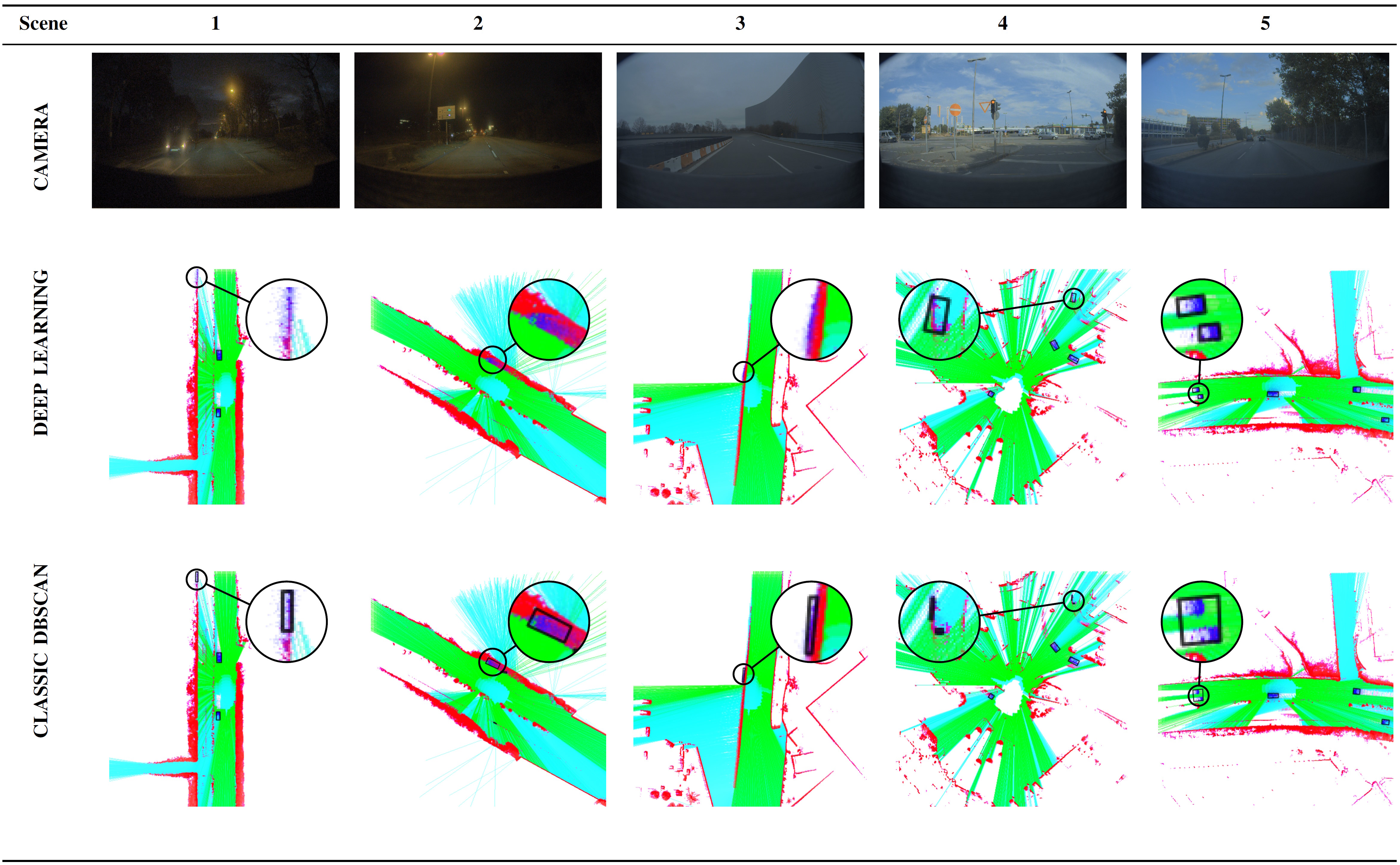}}}
		\caption{Qualitative comparison of the classic DBSCAN and our deep learning-based approach with each column representing a different scene. Camera reference images are shown on top, our deep learning-based rotated bounding box object detection results overlaid on the dynamic grids in the middle, and the classic DBSCAN object detections in the last row. The circular areas are enlarged for better visual comparison. The proposed deep generic dynamic object detector outperforms the classic method in various situations.}
		\label{fig:table_of_figures}
	\end{figure*} 
For the classic method, a precision of $0.51$ and a recall of $0.67$ is achieved. The precision/recall curve for the deep learning-based approach was created by changing the score threshold between $0$ and $1$ based on data 1 and 3 of the test set. At the same recall $0.67$ as the classic method, a precision of $0.926$ is achieved with a score threshold equal to $0.978$. The detection performance of ReDet is thus significantly better than that of the classic method. A remarkably low amount of training data is necessary to learn the generic dynamic object detector, which can be explained by the fact that the dynamic grid already does the heavy lifting of deciding which parts of the environment are static or dynamic. Based on the precision/recall curve with the whole test set, we select the score threshold as $0.75$ for the generic dynamic object detection task with a precision of $0.90$ and a recall of $0.89$. 

Fig.~\ref{fig:table_of_figures} exemplarily shows the comparison of our deep object detection network with the classic DBSCAN clusterer in five exemplary scenes. The front camera images are shown in the first line for reference with detection results overlaid on the dynamic grids underneath. Scenes 1-3 show three typical scenarios, in which the classic clusterer is prone to obtain false positives. In scene 1, the newly appearing grid cells of the road boundary are falsely estimated to be dynamic. Thus, the classic method falsely detects this part as a dynamic object bounding box as shown in black overlaid on the grid. In scene~2, the swaying bushes next to the street result in high dynamic occupancy masses, which lead to a false positive dynamic object detection of the classic method as well. In scene 3, grid cells of the traffic barrier to the left are falsely detected as a dynamic object with DBSCAN. In contrast, our deep generic dynamic object detection method successfully suppresses the false positive by considering the scene context such as the structure of the static environment and free spaces contained in the grid. Scene 4 shows a busy intersection with multiple moving vehicles. Both methods can generally detect these vehicles. However, the vehicle highlighted by the black circle is detected as two dynamic objects by the classic method while the deep learning-based detector correctly extracts just one vehicle. In scene 5, two vehicles drive close to each other are falsely detected as only one vehicle by the classic DBSCAN. In contrast, the deep learning-based approach correctly detects both vehicles as separate objects. 
 
\section{Summary and Conclusion}\label{sec:conclusion}

We proposed a deep neural network-based approach to detect generic dynamic objects on dynamic grid maps from a moving ego vehicle. The problem is modeled as an oriented bounding box object detection task with ReDet chosen due to its promising detection performance on aerial images. Dynamic grid maps are encoded to RGB images and fed into the detection network. We only manually labeled a small dataset and extended it with measurements without dynamic objects for improved ghost object suppression and auto-labels obtained from the classic clustering method. Experiments on real-world sensor measurements confirmed that such data extensions significantly improve the detection performance and that promising results for the challenging task of generic dynamic object detection are possible even with very little training data. We also quantitatively and qualitatively verified that the deep learning method outperforms the classic method by considering scene context contained in dynamic grids. Further improvements are expected by coupling this learned grid-based object detector with learned inverse sensor models~\cite{Wei2023} in the dynamic grid fusion process itself. 

\bibliographystyle{IEEEtran}

\IEEEtriggeratref{10}

\bibliography{lit}

\end{document}